\documentclass{article}
\usepackage{ijcai24}

\usepackage{times}
\usepackage{soul}
\usepackage{url}
\usepackage[hidelinks]{hyperref}
\usepackage[utf8]{inputenc}
\usepackage[small]{caption}
\usepackage{graphicx}
\usepackage{amsmath}
\usepackage{amsthm}
\usepackage{booktabs}
\usepackage{algorithmic}
\usepackage[switch]{lineno}
\usepackage[ruled,linesnumbered]{algorithm2e}
\usepackage{framed}
\usepackage{amsmath}
\usepackage{float}

\newcommand{\modelname}[1]{{LLM4PG}}

\title{Beyond Human Preferences: Exploring Reinforcement Learning Trajectory Evaluation and Improvement through LLMs}

\author{
Zichao Shen $^1$
\and
Tianchen Zhu$^1$\and
Qingyun Sun$^1$\and
Shiqi Gao$^1$\And
Jianxin Li$^1$
\affiliations
$^1$Beihang University, Beijing, China\\
\emails
\{20231074, catezi, sunqy, gaoshiqi, lijx\}@buaa.edu.cn
}

\begin{document}

\maketitle

\begin{abstract}
    Reinforcement learning (RL) faces challenges in evaluating policy trajectories within intricate game tasks due to the difficulty in designing comprehensive and precise reward functions. 
    This inherent difficulty curtails the broader application of RL within game environments characterized by diverse constraints.
    Preference-based reinforcement learning (PbRL) presents a pioneering framework that capitalizes on human preferences as pivotal reward signals, thereby circumventing the need for meticulous reward engineering.
    However, obtaining preference data from human experts is costly and inefficient, especially under conditions marked by complex constraints. 
    To tackle this challenge, we propose a LLM-enabled automatic preference generation framework named \modelname , which harnesses the capabilities of large language models (LLMs) to abstract trajectories, rank preferences, and reconstruct reward functions to optimize conditioned policies. 
    Experiments on tasks with complex language constraints demonstrated the effectiveness of our LLM-enabled reward functions, accelerating RL convergence and overcoming stagnation caused by slow or absent progress under original reward structures. 
    This approach mitigates the reliance on specialized human knowledge and demonstrates the potential of LLMs to enhance RL's effectiveness in complex environments in the wild. 
\end{abstract}

\section{Introduction}

In recent years, Reinforcement Learning (RL)~\cite{Sutton_Barto_2005} has made significant progress in handling intricate decision-making tasks like game AI~\cite{Silver2016MasteringTG},  autonomous driving~\cite{Isele_Rahimi_Cosgun_Subramanian_Fujimura_2018}, machine control~\cite{Levine_Pastor_Krizhevsky_Ibarz_Quillen_2018}, and power grid dispatching~\cite{Yoon2021WinningTL}. 
The main goal for agents in these tasks is to maximize cumulative rewards over time. 
Nonetheless, a key challenge in RL is designing comprehensive and accurate reward functions, which is crucial for evaluating trajectory value and optimizing policies. 
Reward engineering~\cite{Dewey2014ReinforcementLA} is time-consuming and requires collaboration among experienced human experts who understand task objectives, operational logic, and intricate constraints. 
Notably, even skilled engineers encounter significant time investments in delineating intricate reward structures tailored for tasks characterized by diverse constraints. 
Additionally, reward functions devised by humans can be exploited by agents, particularly evident in policy learning where agents leverage flaws in the reward function to prioritize the accumulation of rewards over task fulfillment. 

Previous research has explored methods for reconstructing reward functions from expert trajectories, for example, Inverse Reinforcement Learning (IRL)~\cite{Ng2000AlgorithmsFI}. 
While IRL exhibits commendable performance in several tasks, surpassing human-level proficiency remains difficult and requires a large number of expert demonstrations, making it impractical for numerous tasks where the provision of demonstrations is arduous. 
In contrast, Preference-based Reinforcement Learning (PbRL)~\cite{Christiano_Leike_Brown_Martic_Legg_Amodei_2017} presents a more adaptable and convenient alternative. 
Here, human experts can easily express preferences for agent trajectories, elucidating desired behaviors synonymous with task objectives. 
Especially in tasks with various intricate constraints, human experts can provide empirical judgments without the necessity for highly abstract formal modeling. 
In PbRL paradigms, the reward function is learned from human feedback on trajectory preferences, guiding agents to achieve specified goals or learn the required behavior. 
Recent studies highlight the benefits of adequately incorporating human preferences in the PbRL framework. 

However, obtaining preference data from human experts is costly and inefficient, especially in scenarios with complex constraints. 
On the one hand, the number of human experts with rich task experience is limited, resulting in inefficient preference data queries. 
On the other hand, complex task constraints make the intrinsic judgment logic of trajectory preferences more complex and prone to errors. 
The key to addressing these challenges lies in synthesizing various information in the environment and efficiently generating accurate trajectory preferences to guide the agent in continuously optimizing its performance. 
Recent advancements in Large Language Models (LLMs)~\cite{Brown_Mann_Ryder_Subbiah_Kaplan_Dhariwal_Neelakantan_Shyam_Sastry_Amanda_et_al._2020,openai2024gpt4} have demonstrated their effectiveness across various domains, showcasing their ability to understand context and generalize across different scenarios. 

To address these challenges, we propose an LLM-enabled automatic preference generation framework named \modelname , which leverages LLMs to abstract trajectories, rank preferences, and reconstruct reward functions for the optimization of conditional policies. 
Specifically, leveraging LLMs facilitates the ranking of preferences for agent trajectories, subsequently informing the training of reward predictors. 
This predictor is integrated with downstream RL algorithms to train the agent. 
Empirical findings underscore the capacity of reward functions reconstructed via LLMs to significantly accelerate the convergence process of RL algorithms, even in scenarios where performance was previously exhibiting slow or near-stagnant performance under the original reward structure. 

The highlights of the proposed \modelname~ are summarized as follows: 
\begin{itemize}
    \item \modelname~ is agnostic to specific reinforcement learning algorithms, leveraging the generalization capabilities of LLMs to address the issue of sparse rewards across various environments.
    \item \modelname~ streamlines the complexity of reward design, allowing direct manipulation of reward functions using natural language, which in turn diminishes the need for extensive expert knowledge. 
    \item Using \modelname, we can articulate special constraints in natural language, enabling large language models to adaptively guide agent training through modulation of the reward function that aligns with the posited constraints. 
\end{itemize}
\section{Related Works}


In addressing the challenge of designing reward functions under complex tasks, numerous approaches have been proposed. These methods encompass reward shaping, Preference-Based Reinforcement Learning, and the utilization of Large Language Models (LLMs) for reward function design.

\noindent\textbf{Reward shaping}~\cite{Ng1999PolicyIU} refers to the practice of introducing additional rewards to encourage agents to optimize toward the desired behaviors. 
For example, RND~\cite{Burda_Edwards_Storkey_Klimov_2018} encourages agents to explore unknown or novel states by adding additional intrinsic rewards through curiosity mechanisms and prediction networks, thus avoiding the problem of local optimality caused by sparse rewards. 
In addition, LIU et al.~\cite{Liu_Luo_Zhong_Chen_Liu_Peng_2019} segmented the trajectory of the original agent and learned from each segment to shape the reward of each part. 
However, designing rewards often requires domain-specific knowledge and poorly designed reward schemes can sometimes have detrimental effects. 
To address these issues, Inverse Reinforcement Learning (IRL) ~\cite{Ng2000AlgorithmsFI} is proposed to learn the reward function for reinforcement learning from optimal sequences of interactions. 

\noindent\textbf{Preference-based reinforcement learning}. 
In Preference-based reinforcement learning (PbRL), humans provide preference labels between the behaviors of two agents, and the agent learns based on this human feedback~\cite{Christiano_Leike_Brown_Martic_Legg_Amodei_2017}. 
Furthermore, Ibarz et al.~\cite{Ibarz2018RewardLF} enhanced the efficiency of this method by incorporating additional forms of feedback. 
The research by Wu et al.~\cite{Wu2021RecursivelySB} and Stiennon et al.~\cite{Stiennon2020LearningTS} further indicates that preference-based learning PbRL approaches can be utilized to fine-tune large language models. 

\noindent\textbf{LLM for reward design}. 
Kwon et al.~\cite{kwon2023reward} demonstrated the effectiveness of LLMs in generating reward signals, surpassing traditional supervised learning methods. 
Wang et al.~\cite{wang2024socially} explored LLMs' role in socially conscious reward design, highlighting their potential for integrating ethical considerations. 
Ma et al.~\cite{Ma2023EurekaHR} proposed using LLMs to generate code for reward functions, showcasing their versatility beyond linguistic tasks. 

\section{Methodology}
\begin{figure*}
    \centering
    \includegraphics[width=\textwidth]{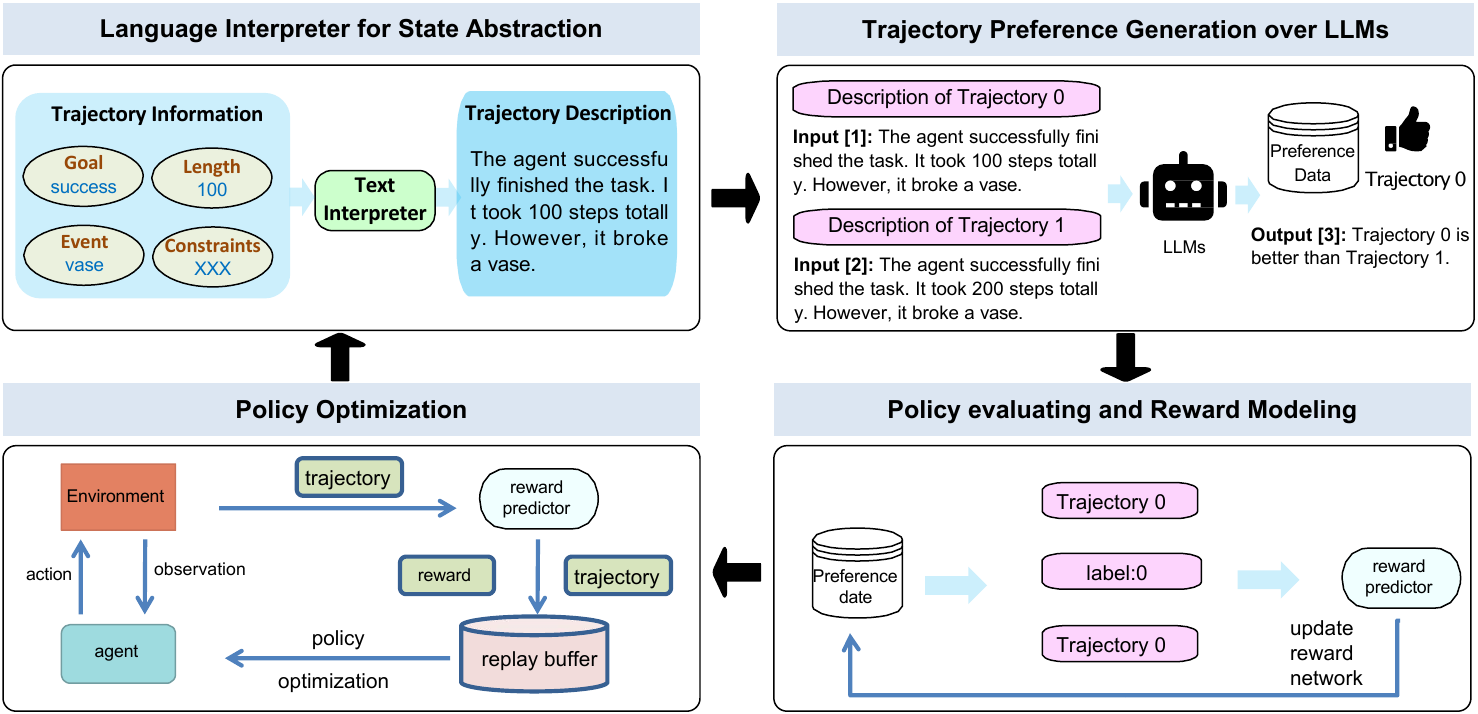}
    \caption{The overall framework of proposed \modelname~.}
    \label{fig-framework}
\end{figure*}

We propose \modelname~, a LLM-enabled automatic preference
generation framework. 
\modelname~ consists of two primary phases, as illustrated in Figure~\ref{fig-framework}, the training phase of the reward predictor, followed by the integration phase with downstream reinforcement learning (RL) algorithms. 
In the first training phase, we feed paired trajectories of agent-environment interactions to a large language model, leveraging the preferences inferred by the language model to train the reward predictor. 
In the second integration phase, we utilize established RL algorithms, in conjunction with the rewards provided by the reward predictor, to further train and derive the final decision-making network model. 
The full procedure of our \modelname~ is summarized in Algorithm~\ref{alg:framework}.
\unskip

\subsection{Language Interpreter for State Abstraction}
Firstly, we propose a Language Interpreter to abstract the state of the agent into a natural language. 
The Language Interpreter is defined as a function $f:S \Rightarrow NL$, where $S$ is the state information of an agent, and $NL$ is the description of the natural language of preference. 
This can be achieved by leveraging multi-modal large models or by establishing specific rules for narration. 
The natural language description should encompass various pieces of information, allowing the LLM to weigh and consider relevant details, thereby fostering a comprehensive and holistic understanding of the agent.

In our approach, to facilitate understanding and processing, we map the entire trajectory of the agent's interactions, based on predefined rules, into natural language. 
Specifically, we further simplify the agent's trajectory information, highlighting whether the ultimate goal has been achieved, the length of the action sequence, key events, and additional constraints. 
The length of the action sequence indicates the efficiency with which the agent completes its task. 
The key events serve a dual purpose that they assist the large language model in assessing the quality of the agent's trajectory and, on the other hand, if negative or unexpected events occur, they signal to the model to downgrade the evaluation of this particular trajectory.

\subsection{Preference Generation by LLM with Complex Constraints}
After that, we use a LLM to measure the reward of the trajectory. Our paper elects to leverage the large language model for ranking the sampled trajectories, thereby reshaping the reward function. 

Upon mapping the interaction data of the agent into natural language, we leverage a LLM for preference determination. 
In this portion, we need to provide the LLM with a brief introduction of the environment, the objectives of the tasks, and the constraints we have imposed. 
First, we present an overarching description of the environment and outline the process through which the agent must navigate to accomplish its tasks. 
This provides the LLM with a general understanding of the tasks and environment. 
Following this, we present our constraints and requirements in the subsequent section, which serve as pivotal reference points for the LLM when making preference decisions. 
Next, we input the natural language descriptions of the two trajectories being compared into the LLM, tasking it with identifying and providing the trajectory it prefers. 
Finally, we restrict the LLM to output only three responses, using the prompt: 
\begin{framed}
If the first is better, you should return 0. otherwise, you should return 1. If they are equal, you should return 2.

Your answer should be structured as follows:

Answer: $\langle$your answer (a number)$\rangle$.
\end{framed}

\subsection{Policy evaluating and Reward Modeling from LLM Feedback}
After obtaining the preference data from LLM, we use these preference data to train a reward predictor. 
The reward function encapsulates the direction in which we want the agent to optimize its behavior. 
If a preference exists between two trajectories:
\begin{align}
    ((S_1^1,A_1^1),(S_2^1,A_2^1),...,(S_k^1,A_k^1))\succ\nonumber \\((S_1^2,A_1^2),(S_2^2,A_2^2), ...,(S_k^2, A_k^2)).
\end{align}

Consequently, the corresponding reward function must necessarily entail:
\begin{align}
R(S_1^1,A_1^1)+R(S_2^1,A_2^1)+...+R(S_k^1,A_k^1)>\nonumber \\R(S_1^2,A_1^2)+R(S_2^2,A_2^2)+...+R(S_k^2,A_k^2).
\end{align}

Therefore, we can utilize the preference data from the LLM to train a reward predictor, which fits the reward function under the guidance of the LLM.

Previously, we used the LLM for preference selection among trajectories. 
In formal terms, each instance of choice can be abstracted as a triplet:$(\sigma^1,\sigma^2,\mu)$, where $\sigma^1$ and $\sigma^2$ represent two trajectories, and $\mu$ signifies the preference indicated. 
Here, when $\mu = (1, 0)$, it signifies a stronger preference for the former trajectory. 
Conversely, $\mu = (0, 1)$ indicates a preference for the latter trajectory. 
Meanwhile, $\mu = (0.5, 0.5)$ implies that the LLM does not exhibit a discernible preference between the two trajectories.

\begin{algorithm}
\caption{\modelname,} 
\label{alg:framework}
Initialize parameters of $Q_\theta$ and $r_{phi}$. 
    
Initialize a dataset of preferences $D \leftarrow \emptyset$. 

\For{m in 1...M}{
    $(\sigma^0, \sigma^1) \sim SAMPLE()$
    
    Translate $\sigma^0,\sigma^1$ into natural language $NL$ using language interpreter. 

    Query the LLM for preference $\mu$. 
    
    Store preference $D \leftarrow D \cup \{\sigma^0, \sigma^1, \mu\}$. 
}
\For{each gradient step}{
    Sample minibatch $\{(\sigma^0, \sigma^1, \mu)_j\}^D_{j=1} \sim D$. 

    Optimize $L^{Reward}$ with respect to $\phi$. 
}

Update the agent parameters $\theta$ with $r_{phi}$. 
    
\end{algorithm}

The subsequent step involves training a reward predictor based on the preference data elicited from the LLM. 
Firstly, we define the probability of preference as in ~\cite{Christiano_Leike_Brown_Martic_Legg_Amodei_2017}. 
The probability that trajectory 1 is preferred over trajectory 2 is given by:
\begin{equation}
    \mathrm{P}(\sigma^1\succ\sigma^2) = \frac{\exp \sum \hat{r}(\mathrm{S}_\mathrm{t}^1,\mathrm{A}_\mathrm{t}^1)}{\exp \sum \hat{r}(\mathrm{S}_\mathrm{t}^1,\mathrm{A}_\mathrm{t}^1)+\exp \sum \hat{r}(\mathrm{S}_\mathrm{t}^2,\mathrm{A}_\mathrm{t}^2)}, 
\end{equation}
where $\hat{r}(\mathrm{S}_t, \mathrm{A}_t)$ denotes the reward predicted at timestep $t$ by the reward predictor. 
Advancing from this, we can employ the cross-entropy loss function to train the model, ensuring its outputs align with the expressed preferences:
\begin{align}
    \mathcal{L} = -\sum_{(\sigma^1,\sigma^2,\mu)}\mu(1)\mathrm{log[P}(\sigma^1\succ\sigma^2)] + \nonumber \\ \mu(2)\mathrm{log[P}(\sigma^2\succ\sigma^1)].
\end{align}

However, directly training with the cross-entropy loss may lead to reward predictions with an undesirably broad range, which is unsuitable for downstream training. 
Consequently, to normalize the scale of reward outputs, we augment the loss function with a squared term of the reward values, aimed at reducing their absolute magnitudes:
\begin{align}
    \mathcal{L} = -\sum_{(\sigma^1,\sigma^2,\mu)}\mu(1)\mathrm{log[P}(\sigma^1\succ\sigma^2)] +\nonumber \\ \mu(2)\mathrm{log[P}(\sigma^2\succ\sigma^1)] +\lambda(r_1^2 + r_2^2).
\end{align}
\subsection{Policy Optimization}
We integrate the rewards provided by the trained reward predictor from the preceding phase with a downstream reinforcement learning algorithm to train the agent. 
Here we employ the Proximal Policy Optimization (PPO)~\cite{Schulman_Wolski_Dhariwal_Radford_Klimov_2017} algorithms for policy optimization. 
Departing from conventional reinforcement learning, in our approach, the agent's rewards are not dispensed by the environment, but are instead conferred by the reward predictor. 
Upon conclusion of each episode, the reward predictor assigns a reward to the agent for its interactions in that round. 

\section{Experiments}
We applied the proposed \modelname~ in MiniGrid~\cite{MinigridMiniworld23}, which is a minimalistic grid world environment library hosting a multitude of classical reinforcement learning environments with sparse rewards and discrete action spaces. 
Within MiniGrid, we selected two specific environments for evaluation: ``MiniGrid-Unlock-v0'' and ``MiniGrid-LavaGapS7-v0''. 
The LLM used in the experiment is Mixtral 8×7B~\cite{jiang2024mixtral} and QWen-max~\cite{bai2023qwen}. 

In experiments involving complex constraints and sparse rewards, the baselines we used were as follows.
\begin{itemize}
    \item \textit{Original reward.} We utilized the original rewards provided by the environment.
    \item \textit{Shaped reward.} 
    In instances of task failure, we administered an artificial negative reward, whose magnitude is set equivalent to the number of steps expended in the given episode. 
\end{itemize}
In the experiments involving additional constraints, the baselines we employed were: 
\begin{itemize}
    \item \textit{Lagrange PPO.} Lagrange PPO introduces constraints as penalizing terms, punishing policy optimizations that breach constraints during optimization.
    \item \textit{PPO.} The agents are trained with Proximal Policy Optimization (PPO), prioritizing reward optimization while intentionally overlooking constraint conditions.
    \item \textit{Random.} The agent performs actions randomly. 
\end{itemize}

\subsection{Experiments on MiniGrid-Unlock-v0}
In this environment, the agent must first acquire a key, and then use the key to unlock and exit through a door. 
The task is successful and ends when the agent successfully opens the door. 
In contrast, the task fails and terminates if the agent does not complete the objective within a designated time frame. 

\subsubsection{State space, action space, and original reward}
\begin{itemize}
\item	\textit{State space}. The observation space is a 7x7 discrete observation space, representing the types of items contained within a 7x7 grid in front of the agent. 
\item	\textit{Action space}. The agent's action space is discrete, comprising six possible actions: \{turn left, turn right, move forward, pick up, drop, toggle (to open doors), done\}. 
Note that the ``done'' action is included but serves no functional purpose. 
\item	\textit{Original reward}. The reward for failure remains at zero, while the reward for success is $1 - 0.9*\frac{step\_count}{max\_steps}$.
\end{itemize}

\subsubsection{Experimental Setup}
For the reward predictor, we opt for a two-layer fully connected neural network, with an input dimension of 3, corresponding to the agent's success, the number of steps taken, and the frequency of dropping keys. 
The prompt provided to the LLM is illustrated in Appendix~\ref{app:unlock}. 
In the subsequent reinforcement learning training, we employed PPO~\cite{Schulman_Wolski_Dhariwal_Radford_Klimov_2017}. 

\subsubsection{Results}
\begin{figure*}
\begin{minipage}{0.49\textwidth}
\centering
\includegraphics[width=0.75\textwidth]{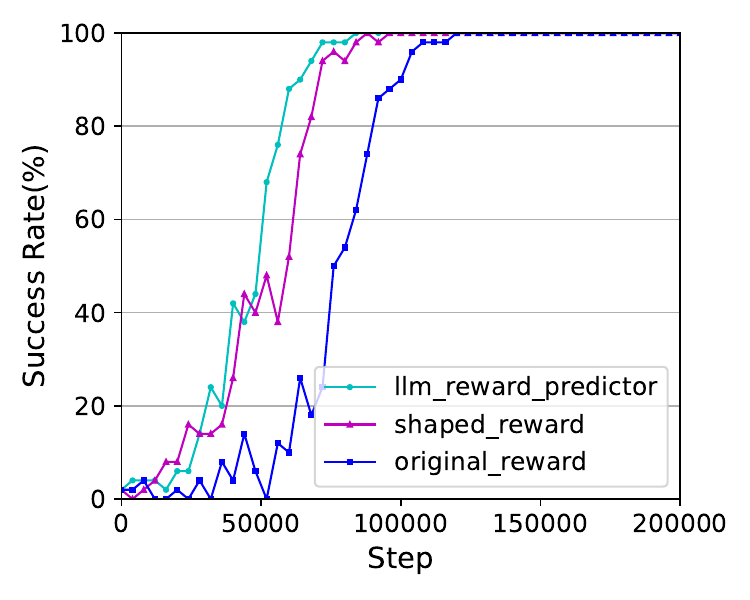}
\caption{Success rate across the training process.}
\label{fig-unlock-train-rate}
\end{minipage}
\begin{minipage}{0.49\textwidth}
    \centering
    \includegraphics[width=0.75\textwidth]{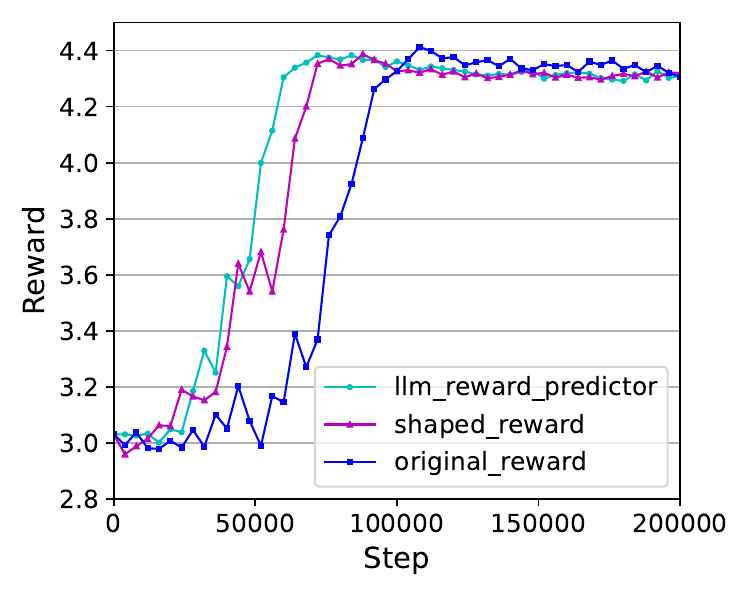}
\caption{The reward across the training process.}
\label{fig-unlock-train-reward}
\end{minipage}
\end{figure*}

Compared to agents trained with the original rewards, those trained using the rewards provided by the reward predictor exhibit a faster convergence rate. 
The agents reach convergence at roughly 50,000 steps during training with the predicted rewards, in comparison, requiring more than twice the number of steps when trained under the original reward system. 
During the training process, the variations in success rates and rewards are illustrated in Figure~\ref{fig-unlock-train-rate} and Figure~\ref{fig-unlock-train-reward}, with the episode rewards compared to the output values of the reward predictor. 

We have also conducted experiments in this environment with added constraint conditions. 
We intend for the agent to attempt to open the door only after dropping the key precisely three times. 
\begin{figure}
\centering
    \includegraphics[width=0.36\textwidth]{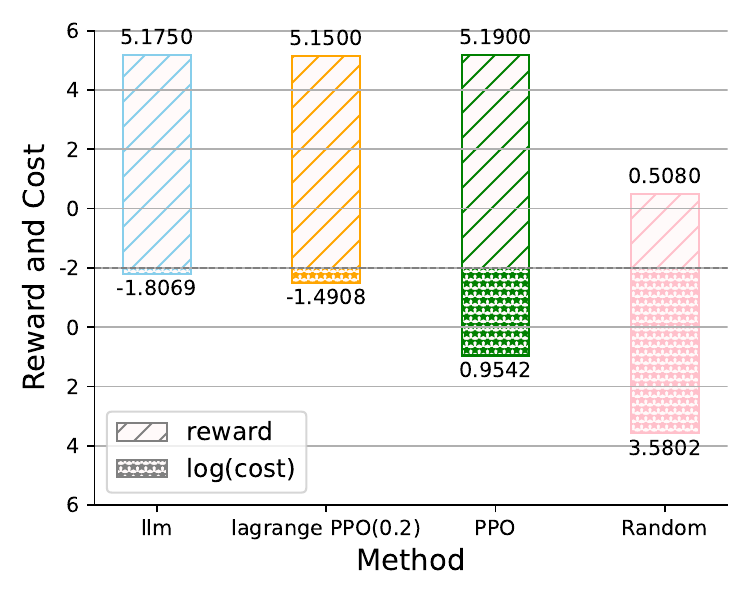}
    \caption{Rewards and costs for each method.}
    \label{fig-llm-lagrange-ppo}
\end{figure}
\begin{figure}
\centering
    \includegraphics[width=0.36\textwidth]{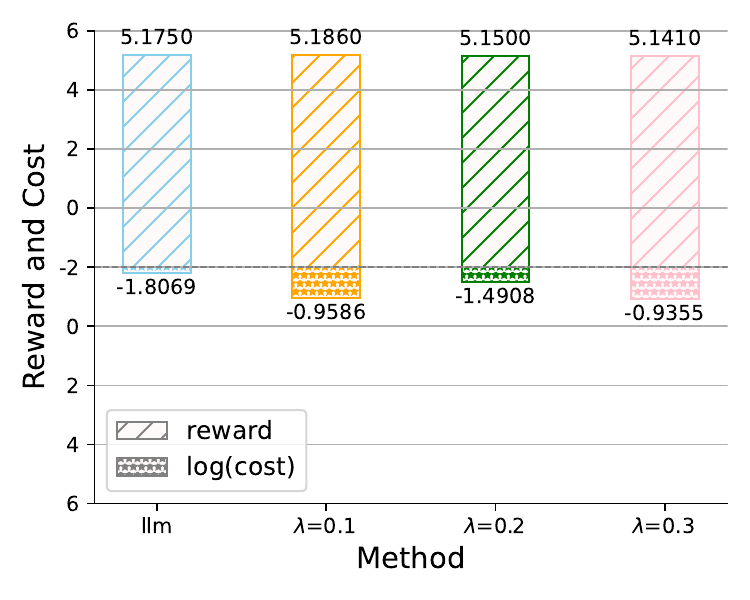}
    \caption{Rewards and costs under different $\lambda$.}
    \label{fig-different-lambda}
\end{figure} 
To convey this intention in the text prompt given to the LLM, we append the following statement: ``We aim for the agent to drop the key exactly three times, with closer to three being more desirable.'' . 
We utilized the LLM to perform preference selection and train the reward predictor, subsequently further training the agent. 
Concerning the measure of cost in the Lagrangian multiplier method, we define the cost as the square of the difference between the number of key drops and three.
The training results are presented in Figure~\ref{fig-llm-lagrange-ppo}. 
In particular, the evaluation of the reward is based on the predictor utilized within the Lagrange PPO. 
Agents trained with reward predictors have attained rewards comparable to those in unconstrained settings, while fulfilling our specified constraint of ideally dropping keys three times per episode. 
Moreover, the Lagrange PPO method has shown notably similar outcomes. 
Nonetheless, the rewards it garners and the costs for breaching constraints are surpassed by our method. 
We have also compared the performance of the Lagrange PPO method under different $\lambda$ values, as illustrated in Figure~\ref{fig-different-lambda}. 
Despite exploring various $\lambda$ values in Lagrange PPO, it remains a challenge to exceed the performance achieved by our method. 



\subsection{Experiments on MiniGrid-LavaGapS7-v0}

This environment exists a lava strip that divides the space into two sections, and the agent must locate and navigate through a singular gap within it. 
The agent is tasked with finding and opening a door on the opposite side to exit. 
This environment challenges the agent's abilities in safe exploration and spatial awareness. 
The game ends in failure if the agent fails to complete the task within a designated time frame or falls into lava. 
Conversely, the game successfully ends when the agent manages to exit through the door. 
\subsubsection{State space, action space, and original reward}
\begin{itemize}
\item	\textit{State space}. The state space is consistent with ``MiniGrid-Unlock-v0''. 
\item	\textit{Action space}. The action space is basically consistent with ``MiniGrid-Unlock-v0''. 
Note that the ``done'', ``pick up'', and ``toggle'' action is included but serves no functional purpose. 
\item	\textit{Original reward}. 
The reward of expiration of time or falling into lava is set to 0 in this environment, while the reward for success is $1 - 0.9*\frac{step\_count}{max\_steps}$.
\end{itemize}

\subsubsection{Experimental Setup}
We construct the reward predictor using a two-layer fully connected neural network with an input dimensionality of 4, representing key state features of the agent's performance: success rate, action duration, lava encounters, and successful lava traversal. 
The prompts for LLMs are detailed in Appendix~\ref{app:lava}. 
For reinforcement learning training, we employ the Proximal Policy Optimization (PPO) algorithm~\cite{Schulman_Wolski_Dhariwal_Radford_Klimov_2017}.

\subsubsection{Results}
\begin{figure}
\centering
    \includegraphics[width=0.36\textwidth]{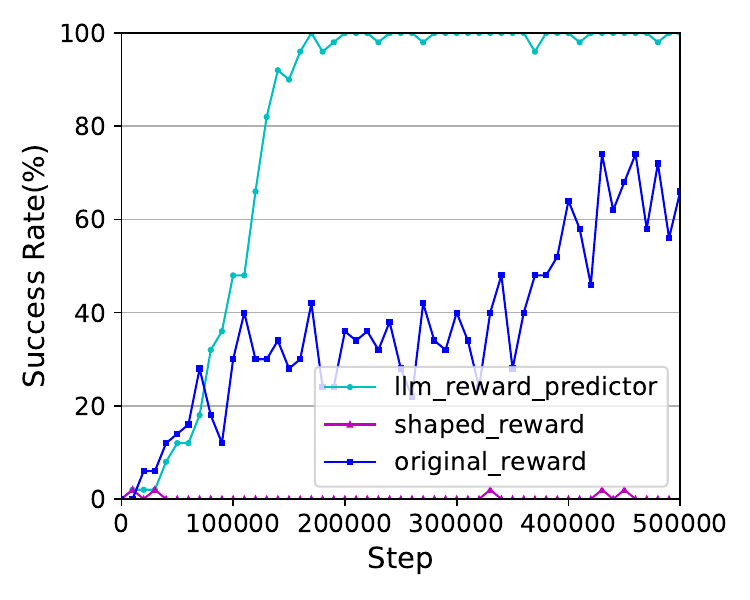}
    \caption{Success Rate across the Training Process.}
    \label{fig-lava-successRate}
\end{figure}

\begin{figure}
\centering
    \includegraphics[width=0.36\textwidth]{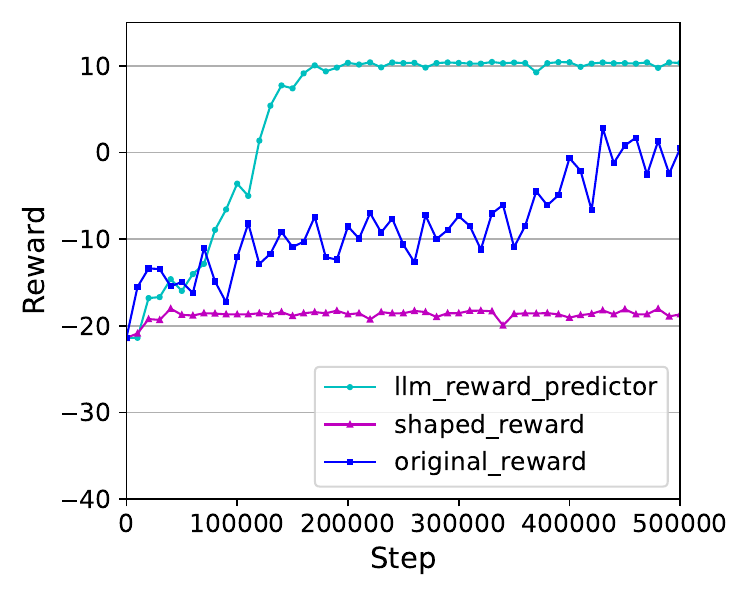}
    \caption{The reward across the Training Process.}
    \label{fig-lava-reward}
\end{figure}

When using the original rewards, the agent training progresses slowly, failing to fully learn how to complete the tasks. 
Conversely, agents trained with our \modelname, exhibit superior performance, whereas those trained with manually crafted reward functions can even fare worse than the original ones. 
The success rates and attained rewards during the training process are presented in Figure~\ref{fig-lava-successRate} and Figure~\ref{fig-lava-reward}. 
\section{Conclusion}


We present \modelname, a novel framework that utilizes large language models (LLMs) to automatically generate trajectory preferences and reconstruct reward models, revolutionizing reinforcement learning for complex environments plagued by complex constraints and reward design.
\modelname~ eliminates the need for domain expertise by allowing the modification of rewards through LLM using natural language trajectory descriptions.
Empirical results highlight \modelname~'s transformative impact, enhancing training speed in existing environments and overcoming conventional reward framework limitations, enabling previously challenging tasks.
Using LLM feedback instead of human feedback, \modelname~ opens new avenues for advancement in reward modeling in RL.

In the future, providing real-time rewards to agents at every step within an episode is essential for tasks that demand rapid responses and the ability to adapt to dynamically changing environments. 
Moreover, incorporating multimodal large models enables the use of direct inputs like images and videos, which is also a direction worthy of further exploration. 

\vspace{6pt}

\appendix
\section[\appendixname~\thesection]{Prompt for LLM}
\subsection[\appendixname~\thesubsection]{The prompt of ``MiniGrid-unlock-v0''}
\label{app:unlock}
\begin{framed}
In a 2 grid world, there is a key and a gate. The agent needs to first get the key and then it can reach the gate and exit. 

We hope that the agent can get the key and reach the gate successfully in the least steps. 

here are two trajectories of the agent. Please choose the better one.

0.<the prompt of a trajectory, for example, the agent successfully got the key and reached the gate. it took 500 steps totally and it dropped the key 67 times.>

1.<the prompt of a trajectory>

please give me which one is better or they are equal.

if the first one is better, you should return 0. otherwise, you should return 1. if they are equal, you should return 2.

Your answer should be structured as follows:

answer:<your answer(a number)>
\end{framed}

\subsection[\appendixname~\thesubsection]{The prompt of ``MiniGrid-LavaGapS7-v0''}
\label{app:lava}
\begin{framed}
In a two-dimensional world, there exists a green exit and a lava strip. The agent is required to first cross the lava strip and then exit through the green exit.

We desire that the agent accomplish this by taking as few steps as possible while avoiding falling into the lava, as doing so would result in its demise.

here are two trajectories of the agent. Please choose the better one.

0. <the prompt of a trajectory, for example, the agent successfully crossed the lava but failed to reach the exit. it took 500 steps totally. it fell into the lava and died..>

1. <the prompt of a trajectory>

please give me which one is better or they are equal.

if the first one is better, you should return 0. otherwise, you should return 1. if they are equal, you should return 2.

Your answer should be structured as follows:

answer:<your answer(a number)>
\end{framed}


\section*{Acknowledgments}

The corresponding author is Qingyun Sun. 
This paper is supported by the National Natural Science Foundation of China (No. 62302023).

\bibliographystyle{named}
\bibliography{ijcai24}

\begin{thebibliography}{}

\bibitem[\protect\citeauthoryear{Achiam \bgroup \em et al.\egroup }{2023}]{openai2024gpt4}
Josh Achiam, Steven Adler, Sandhini Agarwal, Lama Ahmad, Ilge Akkaya, Florencia~Leoni Aleman, Diogo Almeida, Janko Altenschmidt, Sam Altman, Shyamal Anadkat, et~al.
\newblock Gpt-4 technical report.
\newblock {\em arXiv preprint arXiv:2303.08774}, 2023.

\bibitem[\protect\citeauthoryear{Bai \bgroup \em et al.\egroup }{2023}]{bai2023qwen}
Jinze Bai, Shuai Bai, Yunfei Chu, Zeyu Cui, Kai Dang, Xiaodong Deng, Yang Fan, Wenbin Ge, Yu~Han, Fei Huang, et~al.
\newblock Qwen technical report.
\newblock {\em arXiv preprint arXiv:2309.16609}, 2023.

\bibitem[\protect\citeauthoryear{Brown \bgroup \em et al.\egroup }{2020}]{Brown_Mann_Ryder_Subbiah_Kaplan_Dhariwal_Neelakantan_Shyam_Sastry_Amanda_et_al._2020}
Tom Brown, Benjamin Mann, Nick Ryder, Melanie Subbiah, Jared~D Kaplan, Prafulla Dhariwal, Arvind Neelakantan, Pranav Shyam, Girish Sastry, Amanda Askell, et~al.
\newblock Language models are few-shot learners.
\newblock {\em Advances in neural information processing systems}, 33:1877--1901, 2020.

\bibitem[\protect\citeauthoryear{Burda \bgroup \em et al.\egroup }{2018}]{Burda_Edwards_Storkey_Klimov_2018}
Yuri Burda, Harrison Edwards, Amos Storkey, and Oleg Klimov.
\newblock Exploration by random network distillation.
\newblock {\em arXiv preprint arXiv:1810.12894}, 2018.

\bibitem[\protect\citeauthoryear{Chevalier-Boisvert \bgroup \em et al.\egroup }{2024}]{MinigridMiniworld23}
Maxime Chevalier-Boisvert, Bolun Dai, Mark Towers, Rodrigo Perez-Vicente, Lucas Willems, Salem Lahlou, Suman Pal, Pablo~Samuel Castro, and Jordan Terry.
\newblock Minigrid \& miniworld: Modular \& customizable reinforcement learning environments for goal-oriented tasks.
\newblock {\em Advances in Neural Information Processing Systems}, 36, 2024.

\bibitem[\protect\citeauthoryear{Christiano \bgroup \em et al.\egroup }{2017}]{Christiano_Leike_Brown_Martic_Legg_Amodei_2017}
Paul~F Christiano, Jan Leike, Tom Brown, Miljan Martic, Shane Legg, and Dario Amodei.
\newblock Deep reinforcement learning from human preferences.
\newblock {\em Advances in neural information processing systems}, 30, 2017.

\bibitem[\protect\citeauthoryear{Dewey}{2014}]{Dewey2014ReinforcementLA}
Daniel Dewey.
\newblock Reinforcement learning and the reward engineering principle.
\newblock In {\em 2014 AAAI Spring Symposium Series}, 2014.

\bibitem[\protect\citeauthoryear{Ibarz \bgroup \em et al.\egroup }{2018}]{Ibarz2018RewardLF}
Borja Ibarz, Jan Leike, Tobias Pohlen, Geoffrey Irving, Shane Legg, and Dario Amodei.
\newblock Reward learning from human preferences and demonstrations in atari.
\newblock {\em Advances in neural information processing systems}, 31, 2018.

\bibitem[\protect\citeauthoryear{Isele \bgroup \em et al.\egroup }{2018}]{Isele_Rahimi_Cosgun_Subramanian_Fujimura_2018}
David Isele, Reza Rahimi, Akansel Cosgun, Kaushik Subramanian, and Kikuo Fujimura.
\newblock Navigating occluded intersections with autonomous vehicles using deep reinforcement learning.
\newblock In {\em 2018 IEEE international conference on robotics and automation (ICRA)}, pages 2034--2039. IEEE, 2018.

\bibitem[\protect\citeauthoryear{Jiang \bgroup \em et al.\egroup }{2024}]{jiang2024mixtral}
Albert~Q Jiang, Alexandre Sablayrolles, Antoine Roux, Arthur Mensch, Blanche Savary, Chris Bamford, Devendra~Singh Chaplot, Diego de~las Casas, Emma~Bou Hanna, Florian Bressand, et~al.
\newblock Mixtral of experts.
\newblock {\em arXiv preprint arXiv:2401.04088}, 2024.

\bibitem[\protect\citeauthoryear{Kwon \bgroup \em et al.\egroup }{2023}]{kwon2023reward}
Minae Kwon, Sang~Michael Xie, Kalesha Bullard, and Dorsa Sadigh.
\newblock Reward design with language models.
\newblock {\em arXiv preprint arXiv:2303.00001}, 2023.

\bibitem[\protect\citeauthoryear{Levine \bgroup \em et al.\egroup }{2018}]{Levine_Pastor_Krizhevsky_Ibarz_Quillen_2018}
Sergey Levine, Peter Pastor, Alex Krizhevsky, Julian Ibarz, and Deirdre Quillen.
\newblock Learning hand-eye coordination for robotic grasping with deep learning and large-scale data collection.
\newblock {\em The International journal of robotics research}, 37(4-5):421--436, 2018.

\bibitem[\protect\citeauthoryear{Liu \bgroup \em et al.\egroup }{2019}]{Liu_Luo_Zhong_Chen_Liu_Peng_2019}
Yang Liu, Yunan Luo, Yuanyi Zhong, Xi~Chen, Qiang Liu, and Jian Peng.
\newblock Sequence modeling of temporal credit assignment for episodic reinforcement learning.
\newblock {\em arXiv preprint arXiv:1905.13420}, 2019.

\bibitem[\protect\citeauthoryear{Ma \bgroup \em et al.\egroup }{2023}]{Ma2023EurekaHR}
Yecheng~Jason Ma, William Liang, Guanzhi Wang, De-An Huang, Osbert Bastani, Dinesh Jayaraman, Yuke Zhu, Linxi Fan, and Anima Anandkumar.
\newblock Eureka: Human-level reward design via coding large language models.
\newblock {\em arXiv preprint arXiv:2310.12931}, 2023.

\bibitem[\protect\citeauthoryear{Ng \bgroup \em et al.\egroup }{1999}]{Ng1999PolicyIU}
Andrew~Y Ng, Daishi Harada, and Stuart Russell.
\newblock Policy invariance under reward transformations: Theory and application to reward shaping.
\newblock In {\em Icml}, volume~99, pages 278--287, 1999.

\bibitem[\protect\citeauthoryear{Ng \bgroup \em et al.\egroup }{2000}]{Ng2000AlgorithmsFI}
Andrew~Y Ng, Stuart Russell, et~al.
\newblock Algorithms for inverse reinforcement learning.
\newblock In {\em Icml}, volume~1, page~2, 2000.

\bibitem[\protect\citeauthoryear{Schulman \bgroup \em et al.\egroup }{2017}]{Schulman_Wolski_Dhariwal_Radford_Klimov_2017}
John Schulman, Filip Wolski, Prafulla Dhariwal, Alec Radford, and Oleg Klimov.
\newblock Proximal policy optimization algorithms.
\newblock {\em arXiv preprint arXiv:1707.06347}, 2017.

\bibitem[\protect\citeauthoryear{Silver \bgroup \em et al.\egroup }{2016}]{Silver2016MasteringTG}
David Silver, Aja Huang, Chris~J Maddison, Arthur Guez, Laurent Sifre, George Van Den~Driessche, Julian Schrittwieser, Ioannis Antonoglou, Veda Panneershelvam, Marc Lanctot, et~al.
\newblock Mastering the game of go with deep neural networks and tree search.
\newblock {\em nature}, 529(7587):484--489, 2016.

\bibitem[\protect\citeauthoryear{Stiennon \bgroup \em et al.\egroup }{2020}]{Stiennon2020LearningTS}
Nisan Stiennon, Long Ouyang, Jeffrey Wu, Daniel Ziegler, Ryan Lowe, Chelsea Voss, Alec Radford, Dario Amodei, and Paul~F Christiano.
\newblock Learning to summarize with human feedback.
\newblock {\em Advances in Neural Information Processing Systems}, 33:3008--3021, 2020.

\bibitem[\protect\citeauthoryear{Sutton and Barto}{1999}]{Sutton_Barto_2005}
Richard~S Sutton and Andrew~G Barto.
\newblock Reinforcement learning: An introduction.
\newblock {\em Robotica}, 17(2):229--235, 1999.

\bibitem[\protect\citeauthoryear{Wang}{2024}]{wang2024socially}
Zhaoyue Wang.
\newblock Towards socially and morally aware rl agent: Reward design with llm.
\newblock {\em arXiv preprint arXiv:2401.12459}, 2024.

\bibitem[\protect\citeauthoryear{Wu \bgroup \em et al.\egroup }{2021}]{Wu2021RecursivelySB}
Jeff Wu, Long Ouyang, Daniel~M Ziegler, Nisan Stiennon, Ryan Lowe, Jan Leike, and Paul Christiano.
\newblock Recursively summarizing books with human feedback.
\newblock {\em arXiv preprint arXiv:2109.10862}, 2021.

\bibitem[\protect\citeauthoryear{Yoon \bgroup \em et al.\egroup }{2021}]{Yoon2021WinningTL}
Deunsol Yoon, Sunghoon Hong, Byung-Jun Lee, and Kee-Eung Kim.
\newblock Winning the l2rpn challenge: Power grid management via semi-markov afterstate actor-critic.
\newblock In {\em International Conference on Learning Representations}, 2021.

\end{thebibliography}

\end{document}